\pdfoutput=1

\documentclass[11pt]{article}

\usepackage{naacl2021}

\usepackage{times}
\usepackage{latexsym}

\usepackage[T1]{fontenc}

\usepackage[utf8]{inputenc}

\usepackage{microtype}

\title{Applied Language Technology: NLP for the Humanities}

\author{Tuomo Hiippala \\
  Department of Languages, University of Helsinki \\
  P.O. Box 24, Unioninkatu 40 B. \\
  00014 University of Helsinki, Finland \\
  \texttt{tuomo.hiippala@helsinki.fi}}

\begin{document}
\maketitle
\begin{abstract}
This contribution describes a two-course module that seeks to provide humanities majors with a basic understanding of language technology and its applications using Python. The learning materials consist of interactive Jupyter Notebooks and accompanying YouTube videos, which are openly available with a Creative Commons licence.
\end{abstract}

\section{Introduction}

Language technology is increasingly applied in the humanities \citep{hinrichsetal2019}. This contribution describes a two-course module named \emph{Applied Language Technology}, which seeks to provide humanities majors with a basic understanding of language technology and practical skills needed to apply language technology using Python. The module is intended to \emph{empower} the students by showing that language technology is both accessible and applicable to research in the humanities.

\section{Pedagogical Approach}

The learning materials seek to address two major pedagogical challenges. The first challenge concerns terminology: in my experience, the greatest hurdle in teaching language technology to humanities majors is not `technophobia' \citep[480]{ohman2019}, but the technical jargon that acts as a gatekeeper to knowledge in the field \citep[cf.][]{maton2014}. This issue is fundamental to teaching students with no previous experience of programming. To exemplify, beginners in my class occasionally interpret the term `code' in phrases such as ``Write your code here'' as a numerical code needed to unlock an exercise, as opposed to a command written in a programming language. For this reason, the learning materials introduce concepts in Python and language technology in layperson terms and gradually build up the vocabulary needed to advance beyond the learning materials.

The second challenge involves the diversity of the humanities, which covers a broad range of disciplines with different epistemological and methodological standpoints. This results in considerable differences in previous knowledge among the students: linguistics majors may be more likely to be exposed to computational methods and tools than their counterparts majoring in philosophy or art history. Some students may have taken an introductory course in Java or Python, whereas others have never used a command line interface before. To address this issue, the learning materials are based on Jupyter Notebooks, which provide an environment familiar to most students -- a web browser -- for interactive programming. The command line is used for interacting with GitHub, which is used to distribute the learning materials and exercises.

The module also emphasises peer and collaborative learning: 20\% of the course grade is awarded for activity on the course discussion forum hosted on GitHub. All activity -- both asking and answering questions -- counts positively towards the final grade. This allows the students with previous knowledge to help onboard newcomers. According to student feedback, this also fosters a sense of community. The discussion forum is also used to discuss weekly readings, which focus on ethics \citep[e.g.][]{hovyspruit2016, bird2020} and the relationship between language technology and humanities \citep[e.g.][]{kuhn2019, dongetal2020}. These discussions are guided by questions that encourage the students to draw on their disciplinary backgrounds, which exposes them to a wide range of perspectives to language technology and the humanities.

\section{Learning Materials}

The learning materials cover two seven-week courses.

The first course starts by introducing rich, plain and structured text and character encodings, followed by file input/output in Python, common data structures for manipulating textual data and regular expressions. The course then exemplifies basic NLP tasks, such as tokenisation, part-of-speech tagging, syntactic parsing and sentence segmentation by using the spaCy 3.0 natural language processing library \citep{honnibaletal2020} to process examples in the English language. This is followed by an introduction to basic metrics for evaluating the performance of language models. The course concludes with a brief tour of the pandas library for storing and manipulating data \citep{mckinney2010}.

The second course begins with an introduction to processing diverse languages using the Stanza library \citep{qietal2020}, shows how Stanza can interfaced with spaCy, and how the resulting annotations can be searched for linguistic patterns using spaCy. The course then introduces word embeddings to provide the students with a general understanding of this technique and its role in modern NLP, which is also increasingly applied in research on the humanities. The course finishes with an exploration of discourse-level annotations in the Georgetown University Multilayer Corpus  \citep{zeldes2017}, which showcases the CoNLL-U annotation schema.

To what extent the students meet the learning objectives is measured in weekly exercises. The weekly assignments are distributed through GitHub Classroom and automatically graded using \emph{nbgrader}\footnote{\url{https://nbgrader.readthedocs.io}}, which allows generating feedback files with comments that are then pushed back to the student repositories on GitHub. The exercises are also revisited in weekly walkthrough sessions to allow the students to ask questions about the assignments. The students are also required to complete a final assignment for both courses: the first course concludes with a group project that involves preparing a set of data for further analysis, whereas the second course finishes with a longer individual assignment.

All learning materials are openly available with a Creative Commons 4.0 CC-BY licence at the addresses provided in the following section. Access to the weekly exercises is available on request.

\section{Technical Stack}

The learning materials are based on Jupyter Notebooks \citep{kluyveretal2016} hosted in their own GitHub repository.\footnote{\url{https://github.com/Applied-Language-Technology/notebooks}} This repository constitutes a submodule of a separate repository for the website, which is hosted on ReadTheDocs.\footnote{\url{https://applied-language-technology.readthedocs.io}} The notebooks containing the learning materials are rendered into HTML using the Myst-NB parser from the Executable Books project.\footnote{\url{https://executablebooks.org}} This allows keeping the learning materials synchronised, and enables the users to clone the notebooks without the source code for the website. Myst-NB also adds links to Binder \citep{jupyteretal2018} to each notebook on the ReadTheDocs website, which enables anyone to execute and explore the code.

The Jupyter Notebooks provide a familiar environment for interactively exploring Python and the various libraries used, whereas the ReadTheDocs website is meant to be used as a reference work. Both media embed videos from a YouTube channel associated with the courses.\footnote{\url{https://www.youtube.com/c/AppliedLanguageTechnology}} These short explanatory videos exploit the features of the underlying audiovisual medium, such as overlaid arrows, animations and other modes of presentation to explain the topics.

\section{Conclusion}

This contribution has introduced a two-course module that aims to teach humanities majors to apply language technology using Python. Targeted at a student population with diverse disciplinary backgrounds and levels of previous experience, the learning materials use multiple media and layperson terms to build up the vocabulary needed to engage with Python and language technology, complemented by the use of a familiar environment -- a web browser -- for interactive programming using Jupyter Notebooks.

\bibliographystyle{acl_natbib}

\end{document}